\pdfoutput=1

\documentclass[11pt]{article}

\usepackage{EMNLP2023}

\usepackage{times}
\usepackage{latexsym}

\usepackage[T1]{fontenc}

\usepackage[utf8]{inputenc}

\usepackage{microtype}

\usepackage{inconsolata}
\usepackage{multirow}
\usepackage{array}
\usepackage{subfigure}
\usepackage{graphicx}
\usepackage{booktabs}
\usepackage{colortbl} 
\usepackage{xcolor}
\usepackage{color}
\usepackage{mathrsfs} 
\usepackage{amsfonts,amssymb} 
\usepackage{amsmath}
\usepackage{subfigure}
\usepackage{bbm}
\usepackage{verbatim}   
\usepackage{algorithm}
\usepackage{algorithmicx}
\usepackage{algpseudocode}

\DeclareMathOperator*{\argmin}{arg\,min}
\definecolor{tabcl}{RGB}{208,244,222}
\definecolor{lightblue}{RGB}{189,212,230}
\definecolor{lightyellow}{RGB}{255,250,204}
\definecolor{lightgreen}{RGB}{209,240,191}
\definecolor{lightpink}{RGB}{250,217,222}
\renewcommand{\arraystretch}{1.1}

\usepackage{tikz}
 
\newcommand*\halfcirc[1][1ex]{%
\begin{tikzpicture}
\draw[fill] (0,0)-- (90:#1) arc (90:-90:#1) -- cycle ;
\draw (0,0) circle (#1);
\end{tikzpicture}}
\newcommand*\tenpcirc[1][1ex]{%
\begin{tikzpicture}
\draw[fill] (0,0)-- (90:#1) arc (90:54:#1) -- cycle ;
\draw (0,0) circle (#1);
\end{tikzpicture}}
\newcommand*\thirtypcirc[1][1ex]{%
\begin{tikzpicture}
\draw[fill] (0,0)-- (90:#1) arc (90:-18:#1) -- cycle ;
\draw (0,0) circle (#1);
\end{tikzpicture}}
 
\title{Enhancing Contrastive Learning with Noise-Guided Attack: Towards Continual Relation Extraction in the Wild}

\author{\textbf{Ting Wu}$^{1*}$\textbf{,}
        \textbf{Jingyi Liu}$^1$\thanks{\hspace{1mm} Equal Contribution.}\ \textbf{,}
        \textbf{Rui Zheng}$^1$\textbf{,}
        \textbf{Qi Zhang}$^{1,3}$\textbf{,}
        \textbf{Tao Gui}$^2$\textbf{,}
        \textbf{Xuanjing Huang}$^1$ \\
  $^1$School of Computer Science, Fudan University \\
  $^2$Institute of Modern Languages and Linguistics, Fudan University \\
  $^3$Shanghai Key Laboratory of Intelligent Information Processing \\
  {\tt \{tingwu21, liujingyi21\}@m.fudan.edu.cn} 
}

\begin{document}
\maketitle
\begin{abstract}
The principle of continual relation extraction~(CRE) involves adapting to emerging novel relations while preserving od knowledge. While current endeavors in CRE succeed in preserving old knowledge, they tend to fail when exposed to contaminated data streams. We assume this is attributed to their reliance on an artificial hypothesis that the data stream has no annotation errors, which hinders real-world applications for CRE. Considering the ubiquity of noisy labels in real-world datasets, in this paper, we formalize a more practical learning scenario, termed as \textit{noisy-CRE}. Building upon this challenging setting, we develop a noise-resistant contrastive framework named as \textbf{N}oise-guided \textbf{a}ttack in \textbf{C}ontrative \textbf{L}earning~(NaCL) to learn incremental corrupted relations. Compared to direct noise discarding or inaccessible noise relabeling, we present modifying the feature space to match the given noisy labels via attacking can better enrich contrastive representations. Extensive empirical validations highlight that NaCL can achieve consistent performance improvements with increasing noise rates, outperforming state-of-the-art baselines.

\end{abstract}



\section{Introduction}
Alongside the predictive wins of relation extraction~(RE) on various benchmarks~\cite{trisedya-etal-2019-neural,ye-etal-2022-packed}, the need for the ability to acquire sequential experience in dynamic environments stands out the significance. Catering to the real-world learning requirement, a new RE formulation, namely continual relation extraction~(CRE), has been proposed~\cite{wang-etal-2019-sentence}.

Under this topic, catastrophic forgetting~\cite{MCCLOSKEY1989109} where previous knowledge is overwritten as new concepts are learned, remains a key challenge. To prevent forgetting, a variety of sophisticated methods are developed by memory replay~\cite{rebuffi-cvpr2017,sun2020lamal}, weight regularization~\cite{doi:10.1073/pnas.1611835114} or architecture expansion~\cite{hung2019compacting}. \citet{wang-etal-2019-sentence} explicitly store past experiences into a limited memory and replay them to complement new tasks learning. In comparison to exemplars storage, \citet{Dong_Hong_Tao_Chang_Wei_Gong_2021} impose constraints on the update of the important network weights for old knowledge consolidation. As for architecture-based method, it dynamically changes model architectures to acquire new information while remembering previous knowledge~\cite{ehret2021continual}.

\begin{figure}[!t]
\centering
\includegraphics[scale=0.13]{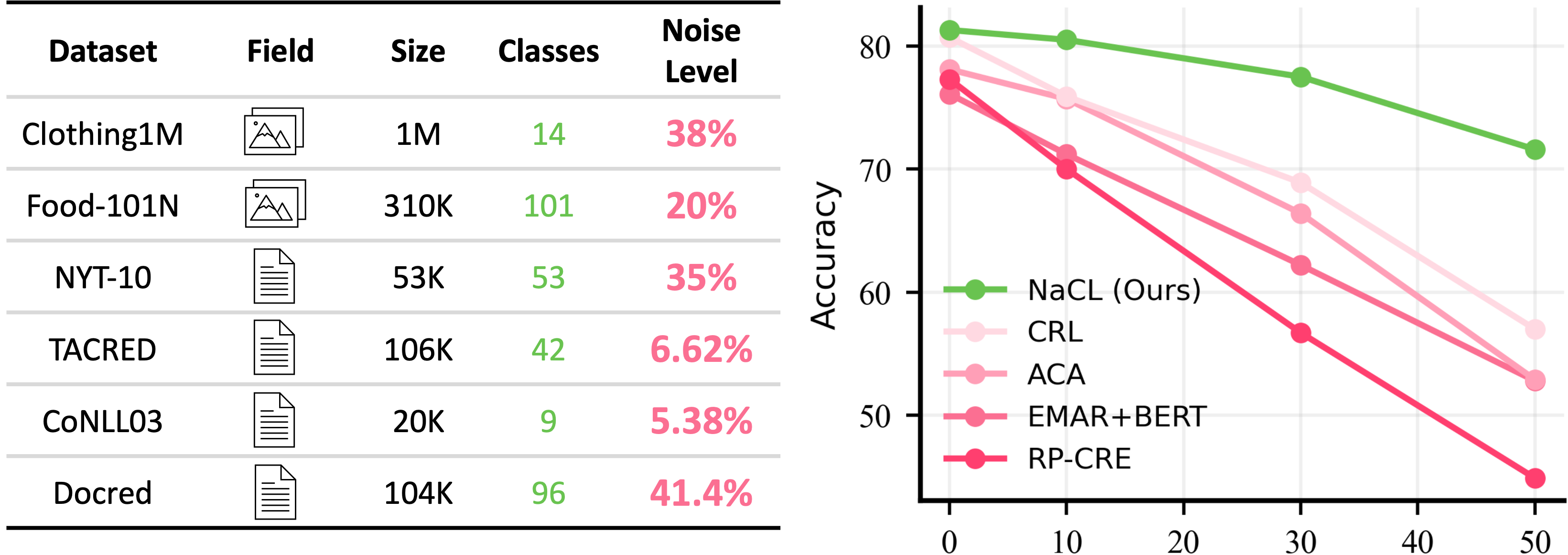}
\caption{Left Table: Noisy labels exist widely in well-annotated benchmarks. Right Plot: Performance of the state-of-the-art CRE methods drop significantly on TACRED with noise ratio ranging from 0\% to 50\%. }
\label{fig:intro}
\end{figure}

Despite the effectiveness, all of these methods implicitly assume the correctness of the labels for the streaming data. In practice, such an assumption is rather artificial even impossible to satisfy since label shifts are inevitable in real-world scenarios. Worse still, official statistics in the table of Figure~\ref{fig:intro} reveal that the widely used benchmarks with elaborate human annotations, likewise, contain a certain proportion of noisy labels. Due to the ignorance of noisy labels over data streams, it is clear to see in Figure~\ref{fig:intro} that state-of-the-art CRE models fail to defend against label inconsistency, resulting in significant performance drops. 

To break the impractical structure of current CRE setup and to enhance the noise-resistant capacity of models, in this paper, we present a more generalized learning setting coined as \textit{noisy-CRE}. In this challenging scenario, there is a potential for mislabeled samples to contaminate the sequential stream in every incremental task. We assume that models trained under the noisy-CRE setting can reflect their ability to adapt to new relations in the real world.

In the face of the great challenge, in this paper, we propose a robust contrastive framework as \textbf{N}oise-guided \textbf{a}ttack \textbf{C}ontrative \textbf{L}earning~(NaCL) for noisy-CRE. Generally, handling noisy labels can be relaxed to a subsequent process of clean sample selection and noisy sample correction. In NaCL, we introduce an auxiliary model to play the two roles. \textbf{First}, at each new task, the auxiliary model will be re-initialized to train for new relations learning. Intriguingly, we term it as \textit{reboot}, which can make the model escape the interference of prior knowledge so that its logit outputs can be a measure of clean sample selection for current task. \textbf{Second}, this model will translate a novel sight into feature space for correction by performing \textit{noise-guided attack}. This attack can actively drive the feature distribution of noisy negatives more aligned with their given labels.

To demonstrate the effectiveness of NaCL, we design two benchmarks based on FewRel and TACRED. Empirical results and in-depth analyses show that our NaCL can achieve consistent improvements when noise rates vary from light to heavy, and it outperforms all state-of-art baselines far ahead. In summary, the contributions of this work are three-fold: 

\indent $\bullet$ We define a practical noisy-CRE setting and construct well-designed benchmarks. To the best of our knowledge, this is the first work to improve the robustness of CRE models against noisy labels. 

\indent $\bullet$ We propose NaCL, a noise-resistant contrastive framework that can jointly prevent catastrophic forgetting and learn with noisy labels. 

\indent $\bullet$ We provide empirical results and extensive assessments to verify the effectiveness of NaCL, outperforming other state-of-the-art baselines adapted from CRE methods by a large margin.

\section{Noisy-CRE Setting Formulation}\label{sec:formulation}

Continual relation extraction is defined as training models on non-stationary data from sequential tasks. In the setup of noisy-CRE, we first define a sequence of tasks $\mathbb{T}=(\mathcal{T}^1,\cdots, \mathcal{T}^n)$. For the $k$-th task $\mathcal{T}^k$, its training dataset is denoted as $\mathcal{D}^k_{\rm train}=\{(x_i, y_i)\}_{i=1}^{N_k}$ containing tuples of the input sample $x_i \in \mathcal{X}$ and corresponding relation label $y_i \in \mathcal{Y}$, where $\mathcal{Y}$ has a probability of rate to be corrupted. Our goal is to train a single model $f_\theta: \mathcal{X} \rightarrow \mathcal{Y}$ parameterized by $\theta$, such that it predicts the label $y=f_\theta(\mathbf{x})\in \mathcal{Y}$ given an unseen test sample $x$ from arbitrary learned tasks.

\begin{figure}[!t]
\centering
\includegraphics[scale=0.15]{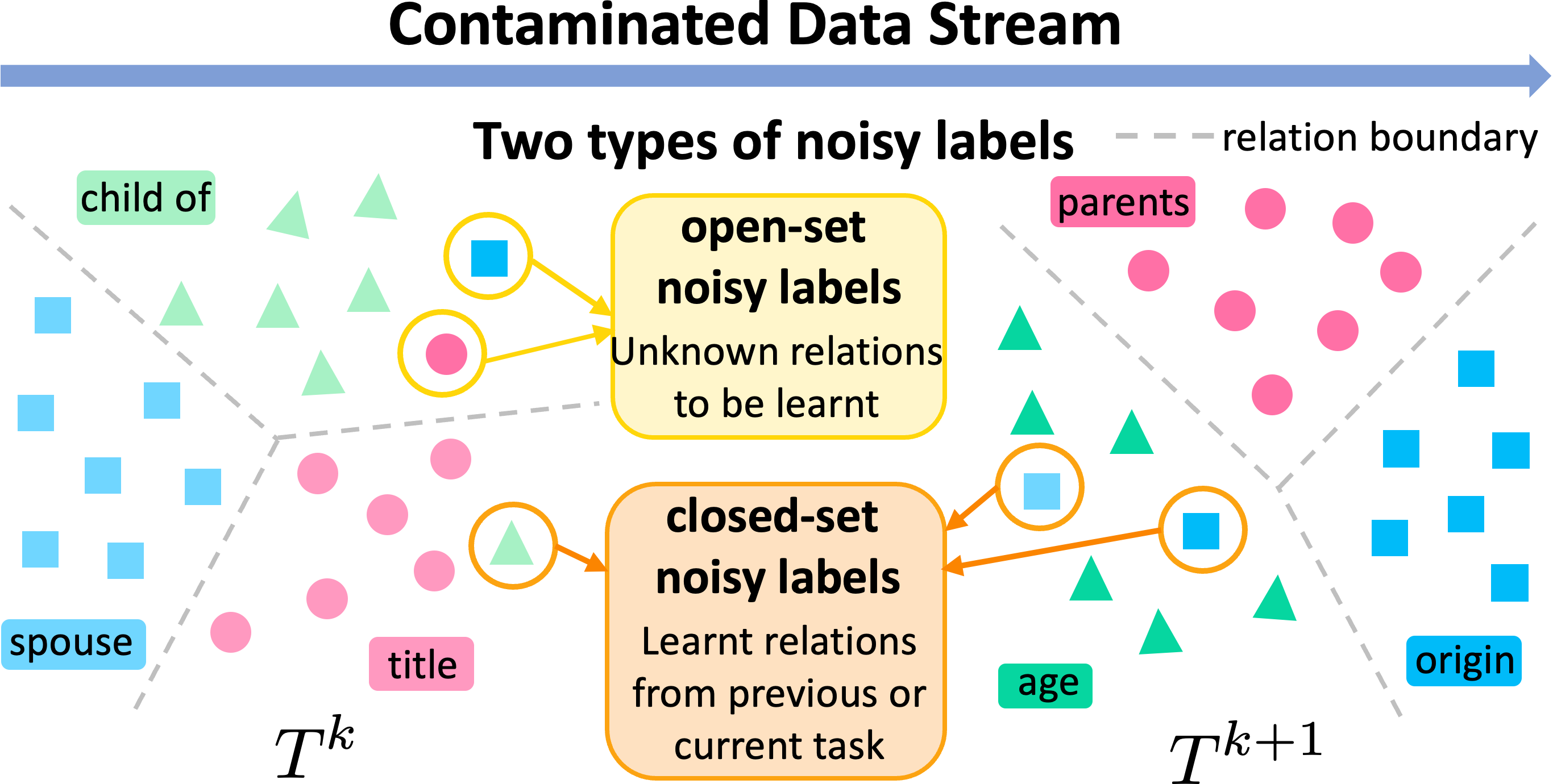}
\caption{The generalized setting of noisy-CRE with two types of noisy labels existing in the contaminated data stream.}
\label{fig:setting}
\end{figure}

\textbf{Protocols for Label Corruption.} In an ideal CRE mode, each task has independent relation space $\mathcal{Y}$. However, for noisy-CRE, due to the inevitable label corruption, this assumption does not hold in the training set. As shown in Figure~\ref{fig:setting}, the relation space $\mathcal{Y}^k$ of the $k$-th task can be contaminated arbitrarily by samples from label space $\mathcal{Y}^i$ with $i\in\{1, \cdots, k-1, k+1,\cdots, n\}$, thus leading to two kinds of noisy labels. When $i\leq k$, we term these noisy labels as \textit{closed-set} ones, since their gold relations are embedded in the model knowledge and can be recovered. In contrast, when $i>k$, the gold relations of the noisy ones are unreachable and formed as \textit{open-set} noise.

\section{NaCL: Towards Noise-resistant CRE}\label{sec:method}
In this section, we present NaCL, our noise-resistant contrastive learning framework designed to simultaneously handle closed-set and open-set noisy labels in the noisy-CRE scenario. 
\subsection{Overall Framework}
\begin{figure}[!t]
\centering
\includegraphics[scale=0.16]{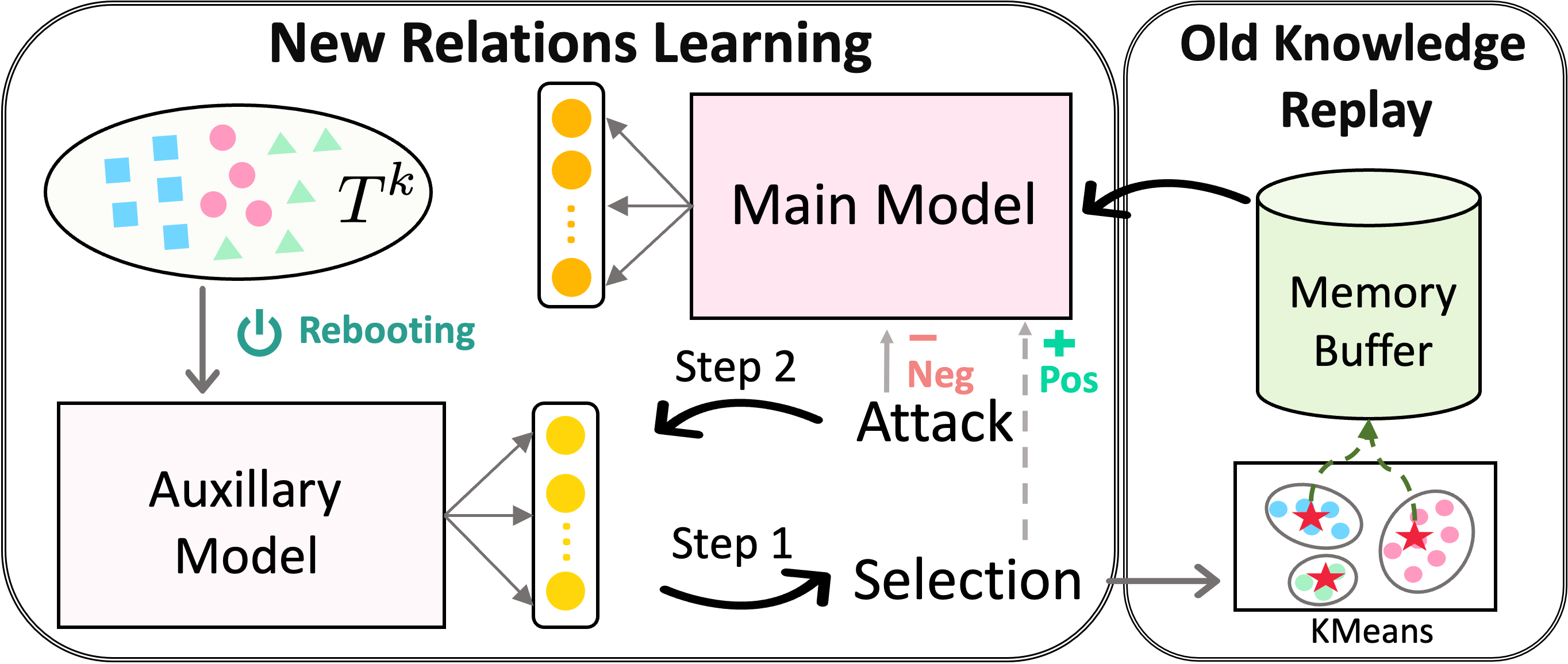}
\caption{Main framework of NaCL and the training pipeline for $\mathcal{T}^k$ learning.}
\label{fig:model}
\end{figure}

Building upon noisy-CRE setting, the learning process of each task contains two components: new relations learning with noisy labels and memory replay for old knowledge consolidation, as presented in the overall framework depicted in Figure~\ref{fig:model}.

\textbf{New Relations Learning.} When learning a new task $\mathcal{T}^k$, the presence of noisy labels can lead to the introduction of false contrastive pairs in vanilla contrastive learning framework. To mitigate this issue, NaCL employs two procedures. First, a rebooted selection process is executed to identify clean positive samples, as described in Section~\ref{sec:purify_positive}. Second, a noise-guided attack is performed on noisy samples to generate hard negatives, which is discussed in Section~\ref{sec:attack_negative}.

\textbf{Old Knowledge Replay.} Once new relations are well-learned at the completion of each task, clean and representative samples stored in the memory buffer will be replayed for old relations prevention.

\subsection{Rebooted Selection for Clean Positives} \label{sec:purify_positive}
To handle the noisy labels, a broadly applied criterion is to select samples with small losses and treat them as clean data. It is inspired by empirical observations that deep learning models tend to learn simple patterns first before overfitting on the noisy labels~\cite{pmlr-v70-arpit17a,zhang2017understanding}.

\begin{figure}[ht]
    \begin{minipage}[ht]{0.48\linewidth}
    \centering
    \includegraphics[scale=0.11]{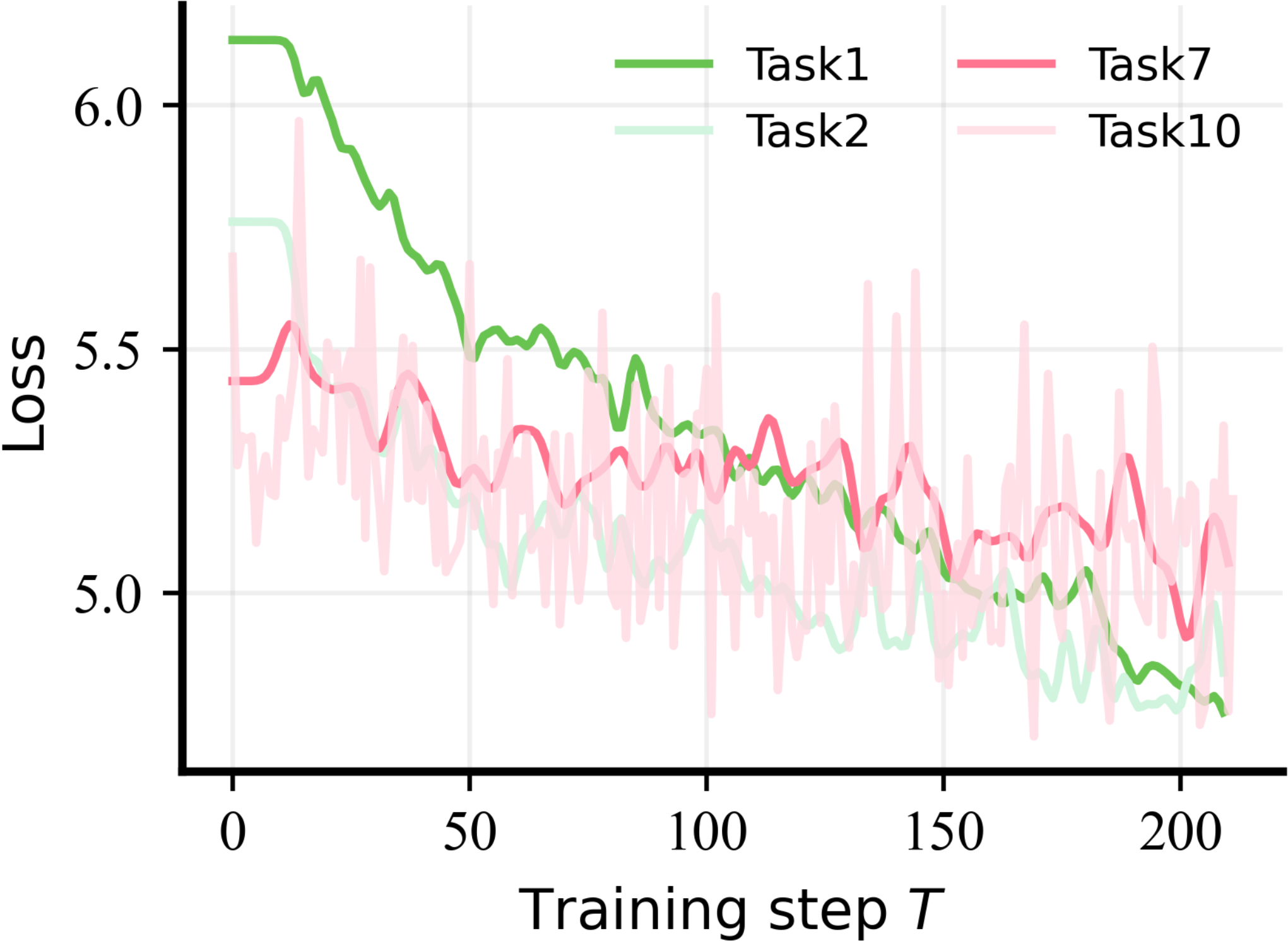}
    \caption{Training loss at different tasks on FewRel with 30\% noise ratio.}
    \label{fig:loss}
    \end{minipage}\hspace{2mm}
    \begin{minipage}[ht]{0.48\linewidth}
    \centering
    \includegraphics[scale=0.105]{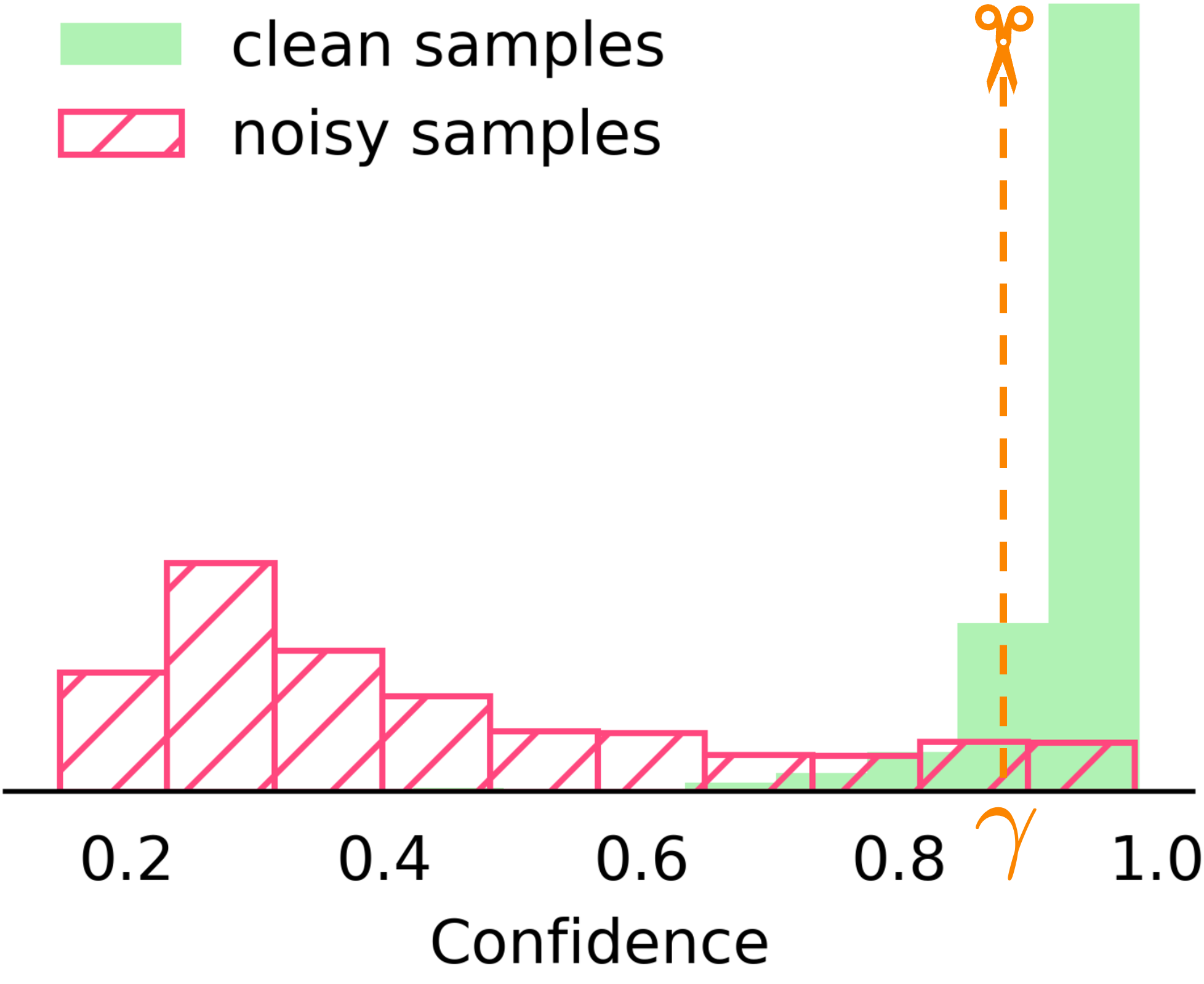}
    \caption{Confidence distribution of clean and noisy samples at Task 10.}
    \label{fig:confidence}
    \end{minipage}
\end{figure}

As shown in Figure~\ref{fig:loss}, we can observe the model quickly converges to a small loss for the first task. However, as the task progresses, an obvious loss threshold between clean and noisy samples gradually disappears. 
We recognize this failure of small-loss-based selection is attributed to the old knowledge of prior tasks embedded in model parameters, which prevents the model from learning incremental tasks from scratch. 

For the sake of overcoming the problem originating from knowledge intervention, we propose to introduce an auxiliary model $f_A(\cdot,\theta^\ast)$ and \textit{reboot} it to help select clean samples at each incremental task. With the decomposition into $f_A = \mathcal{F}_A \circ \mathcal{E}_A$, $\mathcal{E}_A$ being the feature extractor and $\mathcal{F}_A$ the classifier, we train $f_A$ with the following classification loss:
\begin{equation}
    J(\mathbf{x},\mathbf{y}) = -\log p(\mathbf{y}|\mathbf{x})
\label{eq:crossentropy}
\end{equation}

In light of the fact that $f_A(\cdot,\theta^\ast)$ is re-initialized at each new task, it can avoid being intervened by previous knowledge. With a classifier introduced in the auxiliary model $f_A$, we can use the logit probability $p\mathbf{(x)}$ as a measure of confidence to differentiate between clean and noisy samples. As shown in Figure~\ref{fig:confidence}, for the tenth task trained on FewRel with a 30\% noise ratio, a high confidence threshold $\gamma$ successfully identifies almost all clean samples. Consequently, we can predict pseudo clean and noisy set for $\mathcal{T}^k$ as follows:

\begin{equation}
   \mathcal{D}_{train}^k = 
   \begin{cases}
       \widetilde{D}_{\rm clean}(\mathbf{x}), & p(\mathbf{x}) \geq \gamma, \\
       \widetilde{D}_{\rm noisy}(\mathbf{x}), & p(\mathbf{x}) < \gamma, \\
   \end{cases}
   \label{eq:selection}
\end{equation}

\subsection{Noise-guided Attack for Hard Negatives} \label{sec:attack_negative}
Since errors are costly but abstention is manageable, selecting clean samples first and then discarding the noisy ones is a natural approach in the context of learning with noisy labels~\cite{pmlr-v80-jiang18c,xia2022sample}. Nonetheless, over the contaminated data stream, training samples for each task are limited, and thus direct discarding can lead to a loss of abundant context information. Furthermore, the reduction of negative samples will impair contrastive representation learning~\cite{pmlr-v119-chen20j}. Account of the two reasons, making use of noisy samples becomes essential.

\textbf{Noise Correction in Feature Sapce.} One typical way to utilize the noisy samples is to relabel them for correction~\cite{Li2020DivideMix,zhou2021robust}. Faced with the challenge of the co-existence of open-set and closed-set noise, it is impossible for NaCL to apply off-the-shelf techniques to relabel as some noisy labels are unreachable up to current task learning. This inaccessible to label space drives NaCL to translate a novel sight into feature space for noise correction, performed by a variant of targeted attack as \textit{noise-guided attack}.

Noise-guided attack intends to modify the feature to let them match the noisy labels, compared with relabeling that modifies labels to match the given sample features. Within the framework of NaCL, we re-utilize the auxiliary model $f_A$ to implement the attack. As shown in Figure~\ref{fig:model}, at each new task $\mathcal{T}^k$, after training for clean sample selection, $f_A$ will act as the proxy to generate adversarial perturbation on the input embeddings of noisy samples. Assuming the noisy labels $\mathbf{y}$ as the attack targets $\mathbf{y}^{tgt}$, the adversarial loss of $f_A$ is essentially to maximize the probability of classification into $\mathbf{y}^{tgt}$ as follows:

\begin{equation}
    \mathbf{x}^\prime \leftarrow \Pi_\epsilon \big (\mathbf{x} -  \epsilon{\rm sign}(\nabla_{\mathbf{x}^\prime} (J(\mathbf{x}^\prime,\mathbf{y}^{tgt})))\big)
    \label{eq:tgt}
\end{equation}

To further help in generating targeted adversarial examples to match the noisy labels actively, we encourage every adversarial sample to move far away from its starting point in the feature space. To achieve this goal, we add a regularization term to the training objective of Equation~\ref{eq:tgt}:
\begin{equation}
    \begin{aligned}
    \mathbf{x}^\prime \leftarrow \Pi_\epsilon & \big (\mathbf{x} -  \epsilon{\rm sign}(\nabla_{\mathbf{x}^\prime} (J(\mathbf{x}^\prime,\mathbf{y}^{tgt})\\ & + \lambda {\rm KL}(f_A(\mathbf{x};\theta^\ast)||f_A(\mathbf{x}^\prime;\theta^\ast))))\big)
    \label{eq:attack}
    \end{aligned}
\end{equation}
where KL is the Kullback–Leibler divergence, we name this KL regularization as the feature-disruption term, and $\lambda$ is the fixed hyper-parameter to weigh the contribution of this feature disruption.

\textbf{Attack as Hard Negative Mining.} From the perspective of contrastive representation learning, under the noise-guided attack, noisy samples serving as the negatives all move towards the same direction of the feature space where their noisy label lies. To this extent, it can be viewed as hard negative mining which generates more informative negative samples. What's more, given the fixed attack steps $s$, some noisy samples originally closer to the positive region can be successfully pushed into this region for positives diversified. Specifically, denoting the relation-wise centroid as $c_r$ by calculating the mean of the hidden representations for each relation from $\widetilde{\mathcal{D}}_{\rm clean}$, we can obtain $d_{\rm max}$ that measures the maximum euclidean distance of the clean sample to its centroid $c_r$. If the distance between the attacked sample $\mathbf{x^\prime}$ and its corresponding relation centroid $c_r$ is smaller than $d_{\rm max}$, we can recognize this noisy sample is attacked successfully. Consequently, the \textbf{attack success rate}~(ASR) can be calculated as follows:
\begin{equation}
    ASR = \frac{\sum \mathbbm{1} \big[\Vert\mathcal{E}_M(\mathbf{x}^\prime)-c_r\Vert_2 <= d_{\rm max}\big]}{|\widetilde{\mathcal{D}}_{\rm noisy}|}
    \label{eq:asr}
\end{equation}

\textbf{New Contrastive Pool.} We add the successfully attack samples from $\widetilde{\mathcal{D}}_{\rm noisy}$ into the positive set as $\mathcal{D}_{\rm att\mbox{-}pos}$. To this end, we can obtain following contrastive samples pool for current task learning:
\begin{equation}
    A = \underbrace{\widetilde{\mathcal{D}}_{\rm clean} \cup \mathcal{D}_{\rm att\mbox{-}pos}}_{{\rm Positive\ Set}\ P(\mathbf{x})}\cup\ \mathcal{D}_{\rm neg}
\end{equation}

\textbf{Final Learning Objective.} Hence, we come to the training objective of NaCL for new relations learning:
\begin{equation}
    \mathcal{L}_{\rm NaCL} = -\frac{1}{|P(\mathbf{x})|}\sum_{j \in P(\mathbf{x})}\log \frac{\exp\big(\mathbf{z}_i\cdot \mathbf{z}_j /\tau \big)}{\sum\limits_{k \in A} \exp(\mathbf{z}_i\cdot \mathbf{z}_k/\tau)}  
    \label{eq:nacl}
\end{equation}
where $\textbf{z}_\ell={\rm Proj}(\mathcal{E}_M(\mathbf{x}))$, $\tau \in \mathbb{R}^+$ is a scalar temperature parameter.

\subsection{Memory Replay and Inference}
After the stage of $k$-th task training for new relations, NaCL will select representative samples from $\mathcal{D}_{\rm train}^k$ to store in the memory buffer $\mathcal{B}$. The buffer size is the number of memory samples needed for each relation, i.e., 20 in our experiments. Like previous rehearsal-based methods for CRE~\cite{han-etal-2020-continual,cui-etal-2021-refining}, we apply K-Means in the representation space produced by $\mathcal{E}_M$ for exemplar selection, which is only carried out in $\widetilde{\mathcal{D}}_{\rm clean}$. As for each cluster, the sample closest to the cluster center will be selected to store in the buffer $\mathcal{B}$. When the memory buffer is updated with all the seen relations stored, we train $f_M$ with these exemplars of following supervised contrastive loss:
\begin{equation}
    \mathcal{L}_{\rm SCL} = -\frac{1}{|P^\prime(\mathbf{x})|}\sum_{j \in P^\prime(\mathbf{x})}\log \frac{\exp\big(\mathbf{z}_i\cdot \mathbf{z}_j /\tau \big)}{\sum\limits_{k \in \mathcal{B}} \exp(\mathbf{z}_i\cdot \mathbf{z}_k/\tau)}  
    \label{eq:scl}
\end{equation}

\textbf{Relation inference.} Given a test sample $x_i$, nearest class mean~(NCM) is utilized to obtain the relation predicted by $f_M$. Concretely, after the training pipeline of $\mathcal{T}^k$, we can obtain the prototype for each seen relation as $p_r$ by calculating the mean of the features from its corresponding exemplars in the buffer $\mathcal{B}$. To be noted, the calculation of the features is in the space after the projector of the main model $f_M$. Then, we compare the projected representation of $x_i$ with all the prototypes of seen relations and assign the relation label with the closest prototype:
\begin{equation}
   \widetilde{y} = \argmin_{r=1,\dots,C} \Vert {\rm Proj}(\mathcal{E}_M(\mathbf{x})) - p_r\Vert
\end{equation}

\section{Experiments}

\subsection{Experimental Design}
\textbf{Datasets.} We carry out our experiments on widely-used \textbf{FewRel}~\cite{han-etal-2018-fewrel} and \textbf{TACRED}~\cite{zhang-etal-2017-position}. FewRel is an RE dataset that contains 80 relations, each with 700 instances, and TACRED contains 42 relations and 106,264 samples in total. To be noted, previous works for CRE employ two different task partitioning methods to construct the continual benchmarks, one is the imbalanced division based on clustering of relation embeddings~\cite{wang-etal-2019-sentence,han-etal-2020-continual,DBLP:conf/aaai/WuLLHQZX21} and the other is a random partition with balanced relations for each task~\cite{cui-etal-2021-refining,zhao-etal-2022-consistent}. This diversion in task construction makes the baselines incomparable, and we unify them into the same second policy that we split FewRel and TACRED into 10 clusters of relations, leading to 10 tasks and each relation just belongs to only one task. 

\textbf{Noise generation.} We design four levels of random noisy labels to accommodate varying noise rates in real-world data, including clean data, $10\%$ noisy data, $30\%$ noisy data, and $50\%$ noisy data for $\mathcal{D}_{\rm train}^k$ at each task $\mathcal{T}^k$. To generate synthetic noises that contain both close-set and open-set noisy labels, we first randomly flip the relation labels across the whole dataset according to the noise ratio. Then, we partition the dataset based on the flipped relations and cluster them into ten sequential tasks.

\subsection{Baselines} 
We adapt the following state-of-the-art CRE baselines to the proposed noisy-CRE setting and make a comparison with our NaCL model. 

\textbf{EA-EMR}~\cite{wang-etal-2019-sentence} employs memory replay and embedding alignment to tackle the problem of embedding space distortion when training on new tasks.

\textbf{EMAR}~\cite{han-etal-2020-continual} applies episodic memory activation and reconsolidation mechanism to maintain learned knowledge.

\textbf{CML}~\cite{DBLP:conf/aaai/WuLLHQZX21} adopts meta learning and curriculum learning to cope with the challenges of catastrophic forgetting and order-sensitivity in continual relation extraction.

\textbf{RP-CER}~\cite{cui-etal-2021-refining} refines sample embeddings with an attention-based memory network fed with relation prototypes to alleviate catastrophic forgetting.

\textbf{CRL}~\cite{zhao-etal-2022-consistent} proposes a consistent representation learning that maintains the stability of the relation by adopting contrastive learning and knowledge distillation when replaying memory. 

\textbf{ACA}~\cite{wang2022learning} points out catastrophic forgetting problem of previous CRE models mainly lies in shortcuts learning and applies a simple yet effective adversarial class augmentation mechanism to learn more robust representations. 

\textbf{Joint-training} corresponds to training a model from scratch during each incremental task with the total dataset containing all data about new and past classes. We treat the performance of joint-training model on clean dataset as \textit{upper bound}. 

\textbf{Finetuning} in the other hand represents the \textit{lower bound} of performance, as it is a simple training setup that fine-tunes the model at each incremental task with no replay, regularization or model expansion.

\subsection{Training Details and Evaluation Metrics}

\paragraph{Implementation Details.} The main model $f_M$ is composed of a feature extractor $\mathcal{E}_M$ implemented by BERT-base~\cite{devlin-etal-2019-bert} and a projector of 2-layer MLP. For the auxiliary model $f_A$, its feature extractor is implemented by another BERT-base, and the output dimension of the classifier $\mathcal{F}_A$ is the relation numbers of each incremental task, \textit{i.e.}, $8$-dim for FewRel and $4$-dim for TACRED. At each session $k$, we will re-initialized $f_A(;\theta^\ast)$ and train it for 3 epochs to help select the clean samples. Following the baseline methods~\cite{cui-etal-2021-refining,zhao-etal-2022-consistent}, we adopt Adam as the optimizer with the learning rate of 1e-5 on FewRel and 2e-5 on TACRED for both main model and auxiliary model. Considering that baselines all leverage memory replay to help attenuate catastrophic learning, we set a fixed memory size of 20 for relation-wise storage when re-implementing all methods for the sake of a fair comparison. 

\begin{table*}[!t]
\centering
\scalebox{0.835}{\begin{tabular}{p{4.35cm}cccccccccccc}
\toprule
\multirow{3}{*}{\centering \textbf{Models}} & \multicolumn{6}{c}{\textbf{FewRel}} & \multicolumn{6}{c}{\textbf{TACRED}} \\ \cmidrule(r){2-7} \cmidrule(r){8-13}

 & \multicolumn{3}{c}{$Acc$~($\%$)$\ \uparrow$} & \multicolumn{3}{c}{$Forget$~($\%$)$\ \downarrow$} & \multicolumn{3}{c}{$Acc$~($\%$)$\ \uparrow$} & \multicolumn{3}{c}{$Forget$~($\%$)$\ \downarrow$} \\
 
 & \tenpcirc & \thirtypcirc & \halfcirc & \tenpcirc & \thirtypcirc & \halfcirc & \tenpcirc & \thirtypcirc & \halfcirc & \tenpcirc & \thirtypcirc & \halfcirc \\ \midrule
 
Joint-training & 88.1 & 73.7 & 56.4 & -- & -- &-- & 87.3 & 70.2 & 50.4 &-- &-- &-- \\
Finetuning & 10.0 & 9.6 & 9.3 & 100.0 & 100.0 & 100.0 & 12.6 & 12.3 & 11.7 & 100.0 & 100.0 & 100.0\\ 
EA-EMR~\cite{wang-etal-2019-sentence} &22.3&13.5&8.9 &84.3 &93.9 &96.1 &23.6 &17.1 &12.3 &89.5 &95.7 &95.9 \\
EMAR~\cite{han-etal-2020-continual} &37.2 &29.8 &21.2 &64.7 &72.2 &78.2 &19.7 &16.4 &10.3 &78.8 &76.2 &88.5 \\
CML~\cite{DBLP:conf/aaai/WuLLHQZX21} &37.1 &34.0 &25.1 &68.2 &85.3 &89.4 & 22.4 & 20.7 & 18.1 & 70.1 & 79.2 & 81.3 \\
EMAR+BERT &83.0 &77.6 &67.9 &22.1 &33.0 &42.1 &71.2 &62.2 &52.8 &27.7 &37.5 &47.7 \\
RP-CRE~\cite{cui-etal-2021-refining} &77.1 &65.0 &54.2 &30.2 &42.7 &56.7 &70.0 &56.7 &44.9 &37.4 &52.5 &64.7 \\ 
CRL~\cite{zhao-etal-2022-consistent} &77.7&73.0&66.8 &13.7&17.3&19.9 &75.9&68.9&57.0 &21.1&27.4&41.9 \\ 
ACA~\cite{wang2022learning}$\ \dagger$ &84.1 &78.1 &68.3 &18.9 &27.3 &38.9 &75.7 &66.4 &52.9 &25.8 &38.2 &54.6 \\
\midrule
\textbf{NaCL} &\textbf{84.1}&\textbf{83.7}&\textbf{80.5} &\textbf{11.4} & \textbf{16.0} & \textbf{16.8} & \textbf{80.5} & \textbf{77.5} & \textbf{71.6} & \textbf{13.1} & \textbf{16.8} & \textbf{24.6} \\
\bottomrule
\end{tabular}}
\caption{\textbf{Last test accuracy and forgetting} on FewRel and TACRED with noise ratio of $\{\tenpcirc 10\%, \thirtypcirc 30\%, \halfcirc 50\%\}$. We re-implement all the baselines with equal task division and evaluation for a fair comparison. $\dagger$ indicates EMAR+ACA since ACA is implemented based on the backbone of EMAR and RP-CRE, and it achieves better accuracy.} 
\label{tab: main_res}
\end{table*} 

\begin{figure*}[!t]
\centering  
    \subfigure[FewRel $30\%$ noise]{   
    \begin{minipage}[t]{0.245\linewidth}
    \centering  
    \includegraphics[scale=0.103]{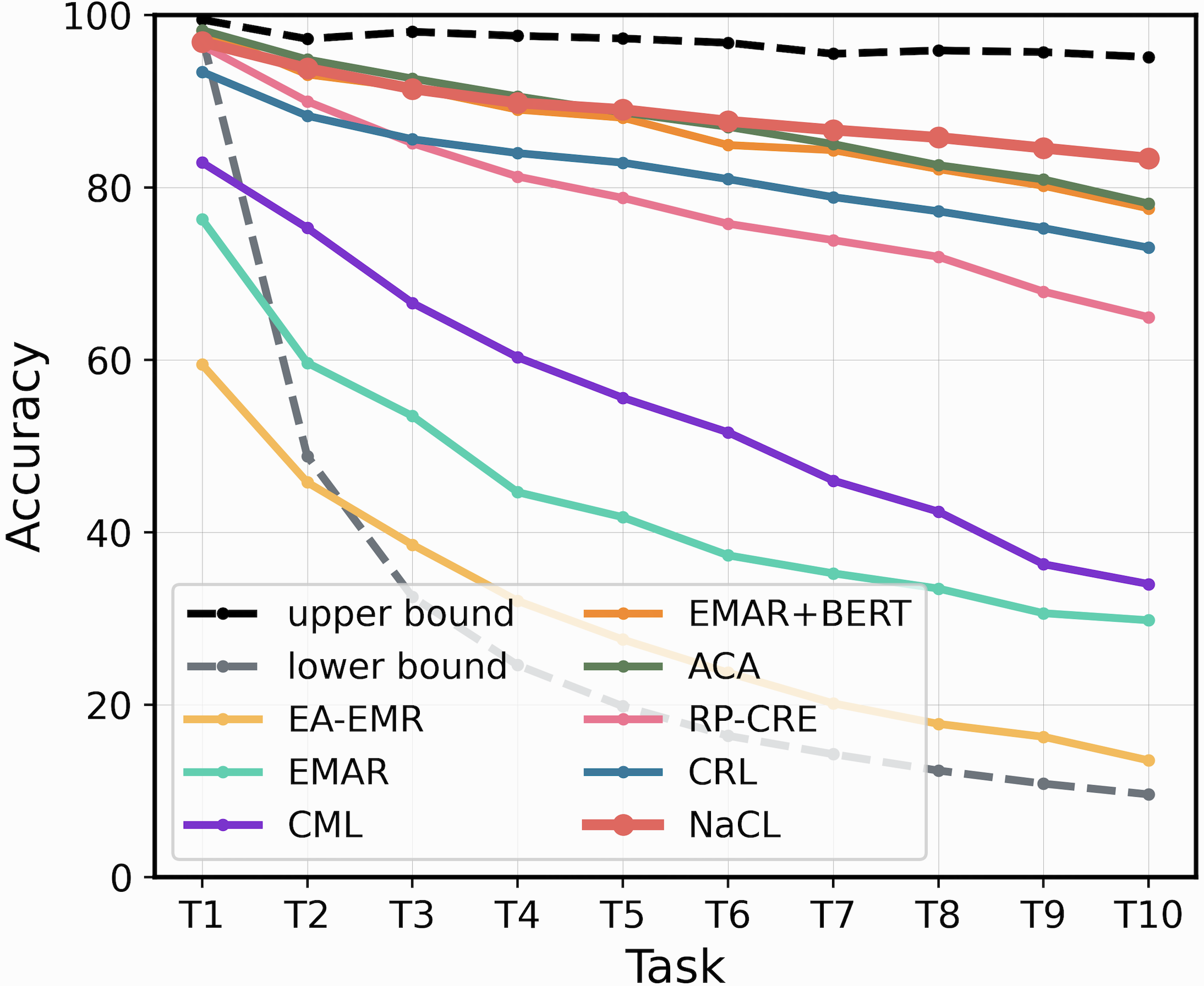}
    \label{fig:fewrel30cl}
    \end{minipage}}
    \subfigure[FewRel $50\%$ noise]{ 
    \begin{minipage}[t]{0.235\linewidth}
    \centering    
    \includegraphics[scale=0.107]{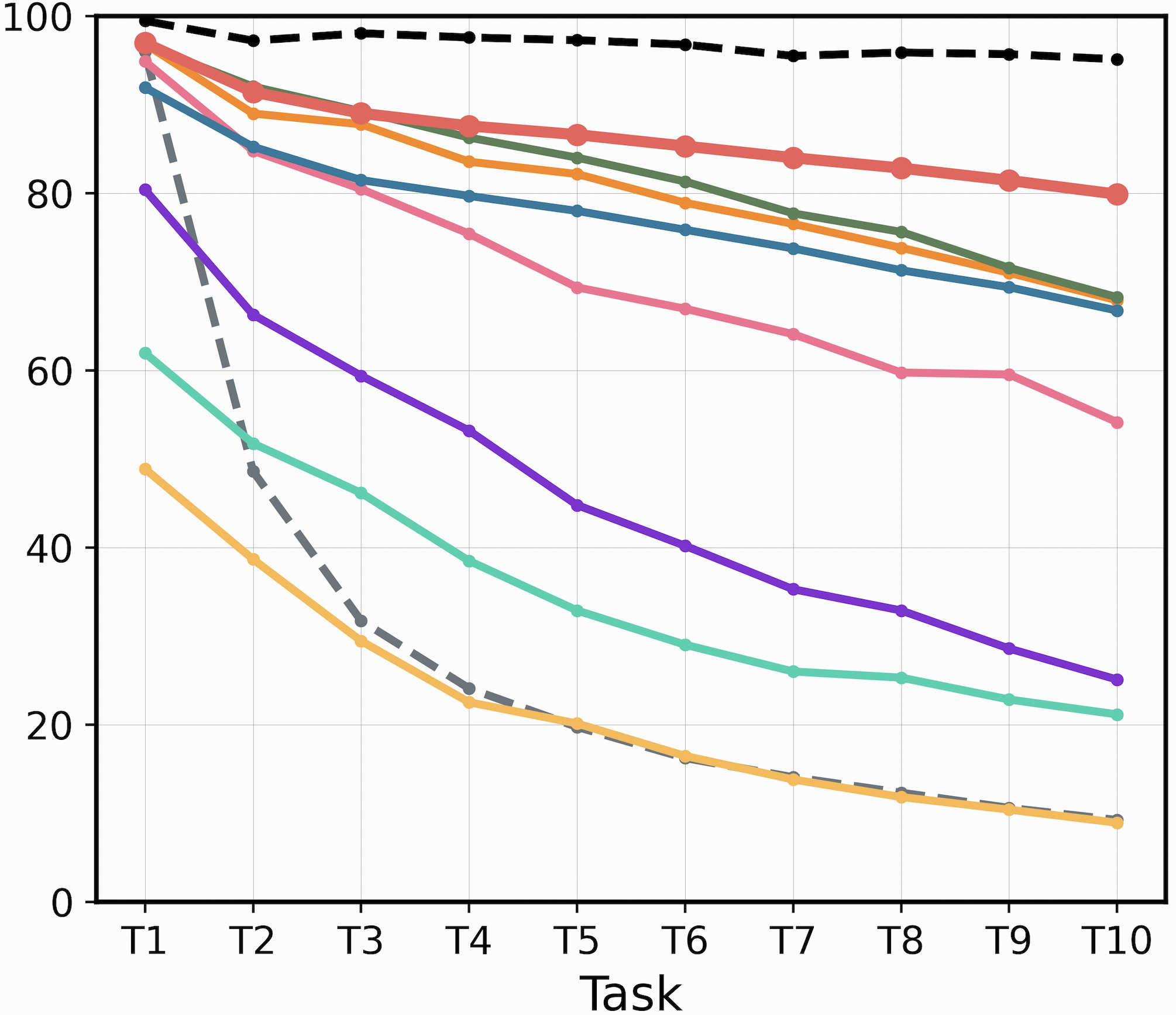}
    \label{fig:fewrel50cl}
    \end{minipage}}
    \subfigure[Tacred $30\%$ noise]{ 
    \begin{minipage}[t]{0.235\linewidth}
    \centering    
    \includegraphics[scale=0.106]{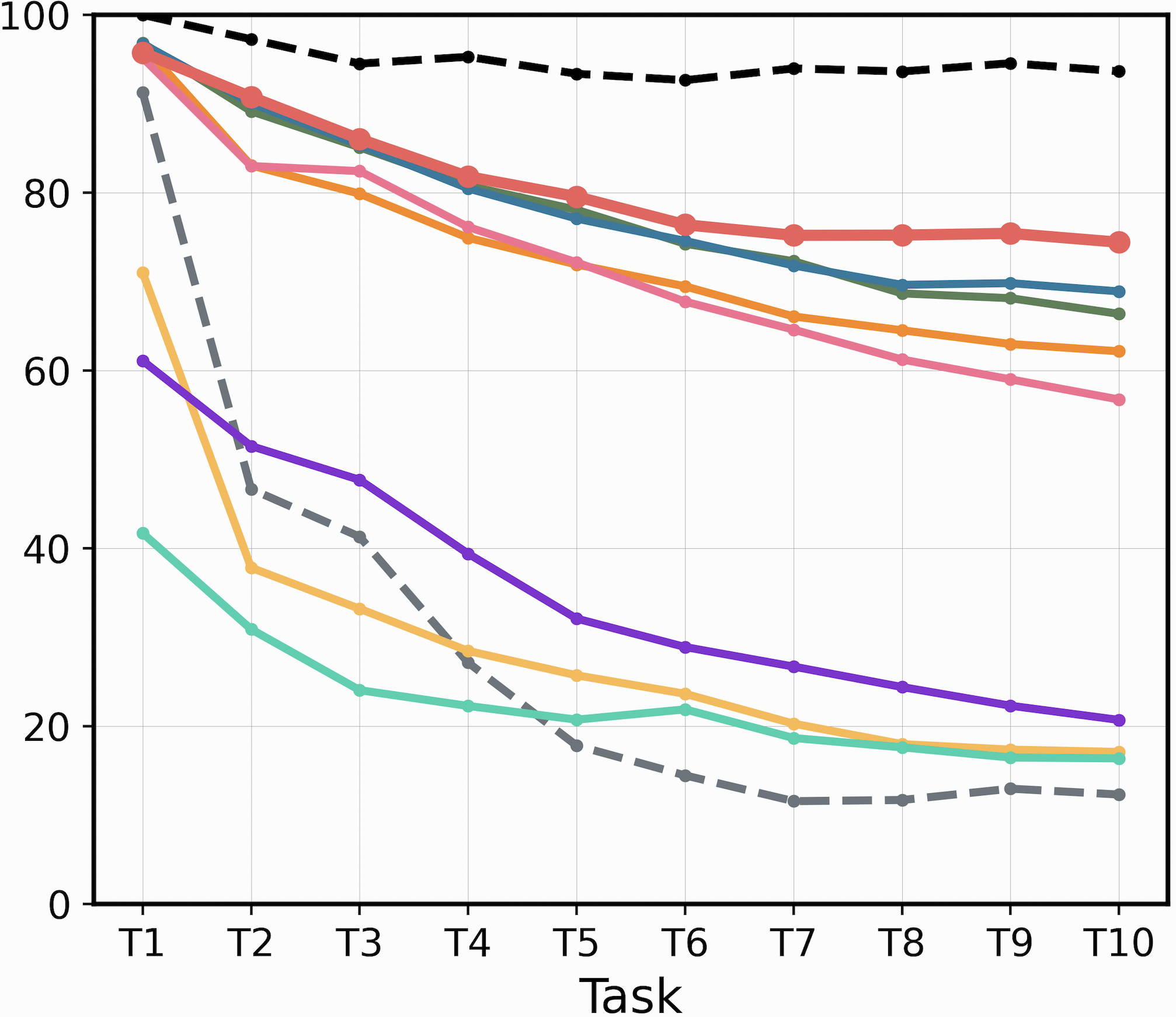}
    \label{fig:tacred30cl}
    \end{minipage}}
    \subfigure[Tacred $50\%$ noise]{ 
    \begin{minipage}[t]{0.235\linewidth}
    \centering  
    \includegraphics[scale=0.109]{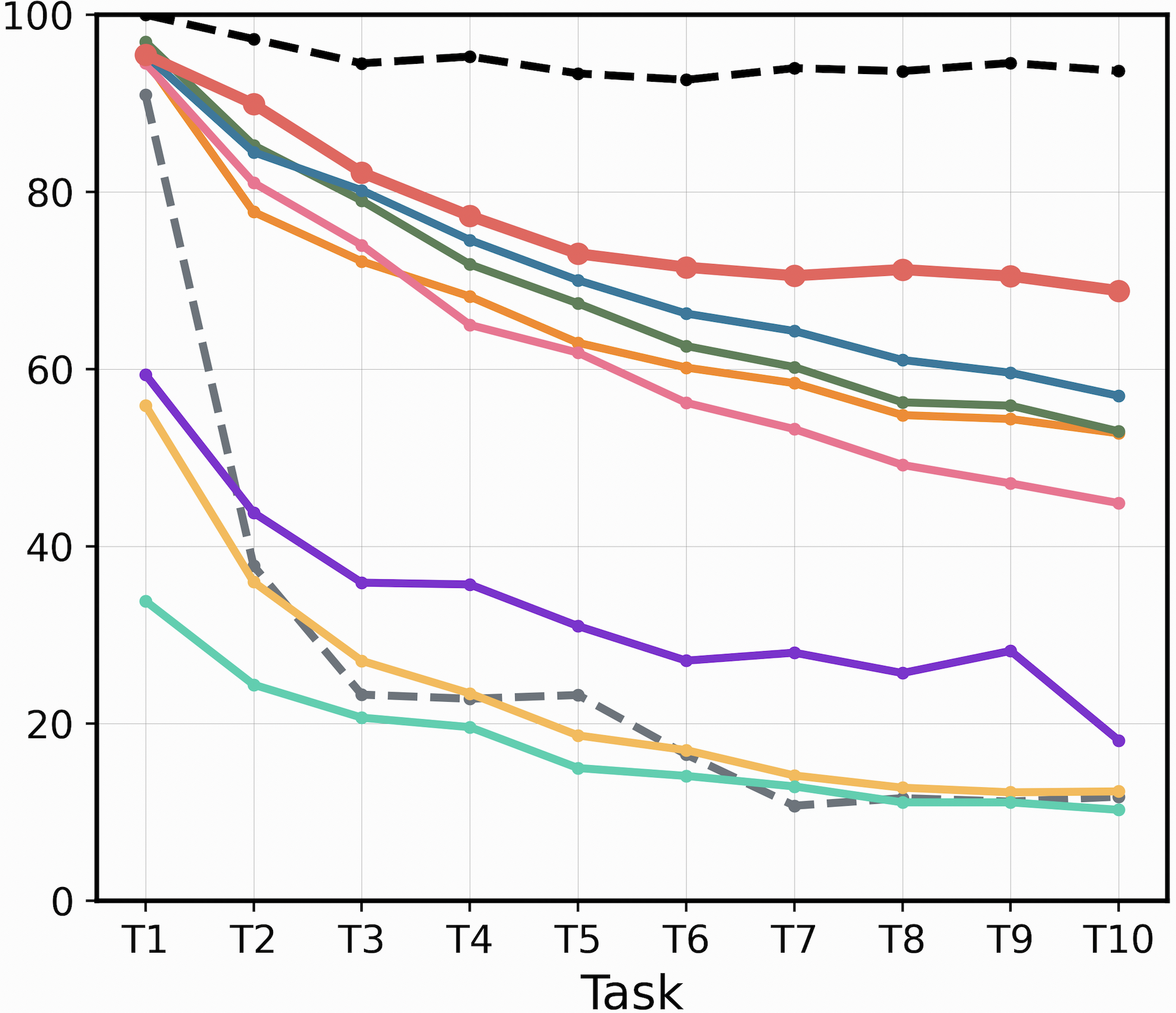}
    \label{fig:tacred50cl}
    \end{minipage}}
\caption{\textbf{Accuracy}~($\%$) \textbf{on all seen relations} at the stage of learning current tasks with varying noise rates on FewRel and TACRED.} 
\label{fig:main_results_samxcl}
\end{figure*}

\textbf{Evaluation Metrics.} As the main performance metric, we adopt \textbf{last test accuracy}, where after all tasks are learned, testing on the test sets of all tasks. We report the average accuracy over 5 random runs. Additionally, we introduce a \textbf{normalized forgetting} metric to quantify the severity of catastrophic learning. As a self-relative metric on the performance drop of the first task, the forgetting measure from previous works~\cite{Liu2020MnemonicsTM} applied to a noisy setting could be misleading since even if a model performs poorly, small forgetting metric values will be observed due to its little information to forget from the beginning. Therefore, we normalize this forgetting on the accuracy of the first task. 
\begin{equation}
        Forget = \frac{|\mathcal{A}_{\mathcal{T}=1}^n - \mathcal{A}_{\mathcal{T}=1}^1|}{\mathcal{A}_{\mathcal{T}=1}^1} 
\label{eq:forget}
\end{equation}
where $\mathcal{A}_{\mathcal{T}=1}^k$ denotes the accuracy on the first task at the session $k$. For \textit{accuracy}, the larger is better, while for \textit{forget}, the smaller will be better.

\subsection{Main Results}
We compare the proposed NaCL with nine baselines on FewRel and TACRED with varying label noise and summarize the results in Table~\ref{tab: main_res}. 

\textbf{Overall Performance.} Table~\ref{tab: main_res} clearly demonstrates that NaCL achieves consistent performance improvements with noise rate from light to heavy, and outperforms all the baselines by a large margin. Furthermore, we can observe that: \textbf{(i)} Apart from our NaCL, all the baselines suffer from the vulnerability of label flips in the continual stream, indicating current CRE models are not resistant to noisy labels. It is apparent to see as the noise rate increases, their last test accuracy declines sharply and the forget rate remains high. \textbf{(ii)} Comparison among the baselines validates that BERT-like pretrained language models are better continual learners since EA-EMR, EMAR, and CML that leverage LSTM as main feature extractor attain worse performances. \textbf{(iii)} There is a close connection between model learning accuracy and the ability to defend against catastrophic forgetting. As shown in Figure~\ref{fig:main_results_samxcl}, test accuracy over ten incremental tasks depicts a vivid trend that if a model achieves high accuracy at each incremental task, its final forget rate tends to retain at a low level.

\textbf{Purity of Memory Buffer.} As rehearsal-based methods served for old knowledge consolidation, the purity of the memory buffer is vital. Therefore, we compare the ratio of clean samples in the memory between NaCL and the high-performing baselines. As shown in Table~\ref{tab:purity}, we observe that EMAR-BERT, RP-CRE and CRL all experience a significant decrease in the purity of the memory buffer as the noise rate increases. In contrast, NaCL is able to maintain comparative purification even with the noise rate increasing. 

\begin{figure}[!t]
\centering
\includegraphics[scale=0.2]{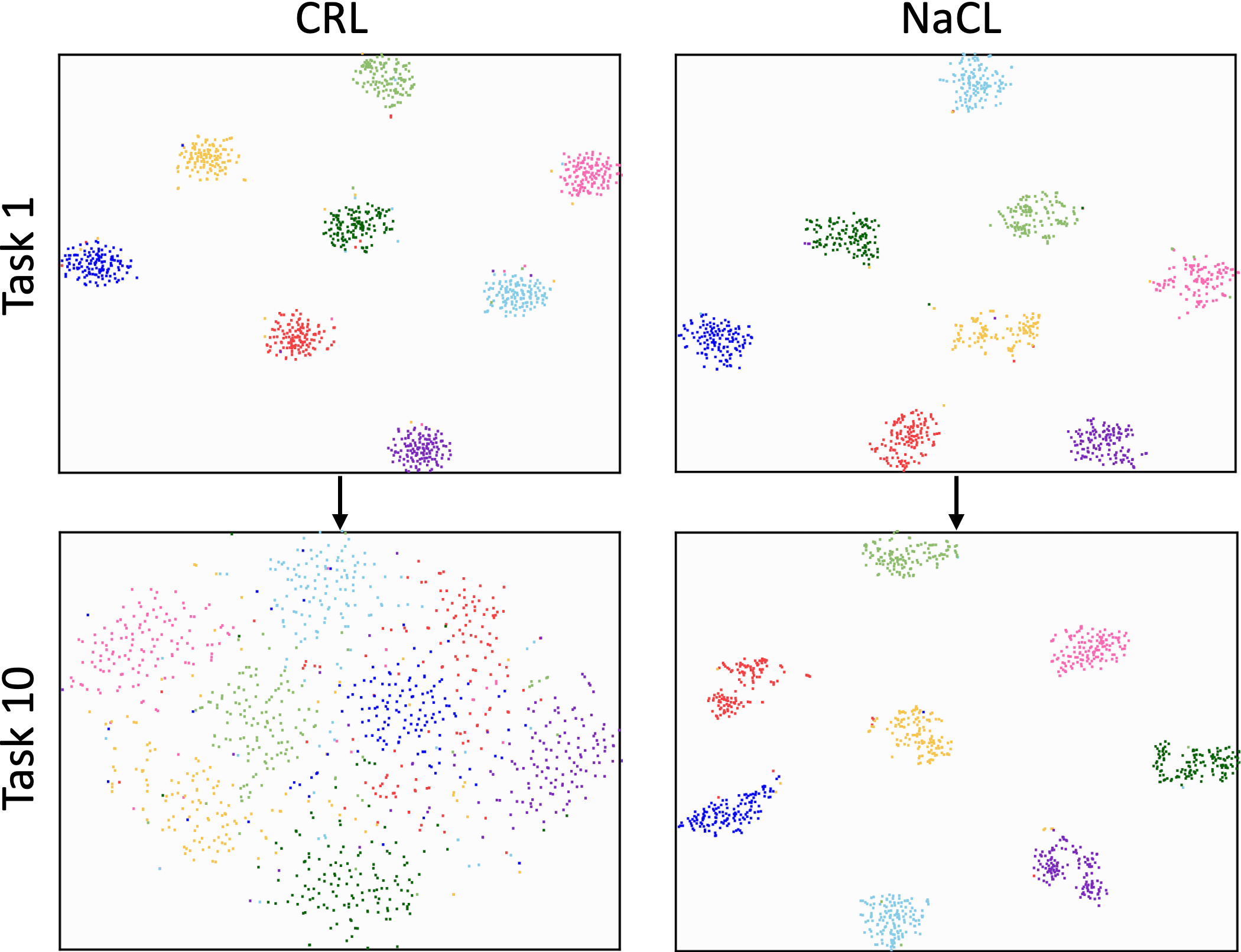}
\caption{\textbf{t-SNE visualization} of relation representation learned from Task 1 and tested by CRL and NaCL at the last task, with a noise rate of $50\%$ on FewRel. Colors stand for different relations.}
\label{fig:tse_vis}
\end{figure}

\begin{table}[ht]
    \centering
    \scalebox{0.83}{\begin{tabular}{ccccccc}\toprule
    & \multicolumn{3}{c}{FewRel} & \multicolumn{3}{c}{TACRED} \\ \cmidrule(r){2-4} \cmidrule(r){5-7}
       noise rate$(\%)$  & 10 & 30 & 50 & 10 & 30 & 50 \\ \cmidrule{1-7}
       EMAR-BERT &80.2 &58.9 &40.7 &76.1 &60.0 &46.1\\
       RP-CRE &88.1 &76.4 &63.8 &79.1 &63.1&50.9 \\
       CRL &68.3 &47.2 &36.3 &71.4 &53.6&41.2 \\
       NaCL & \textbf{98.6} & \textbf{96.4} & \textbf{80.3} & \textbf{94.8} & \textbf{82.4} & \textbf{71.5} \\ \bottomrule
    \end{tabular}
    }
    \caption{\textbf{Purity} of the memory buffer.}
    \label{tab:purity}
\end{table}

\textbf{Preserve of Cluster Relative Positions.} We further demonstreate the t-SNE visualization of the representations learned at the first task and tested at the subsequent tasks in Figure~\ref{fig:tse_vis}. As we can observe, compared to CRL, NaCL can achieve more compact clustering of the representations in the feature space and better preserve the relative positions of each relation cluster. It is worth noting that when approaching the last task, relations learned with CRL become indistinguishable, while NaCL maintains their structures, revealing that NaCL has a better capacity to prevent catastrophic forgetting.

\section{Analysis and Discussion}

\subsection{Effectiveness of Adversarial Attack}

\begin{figure}[htbp]
    \centering
    \subfigure[FewRel]{   
    \begin{minipage}[t]{0.46\linewidth}
    \centering  
    \includegraphics[scale=0.079]{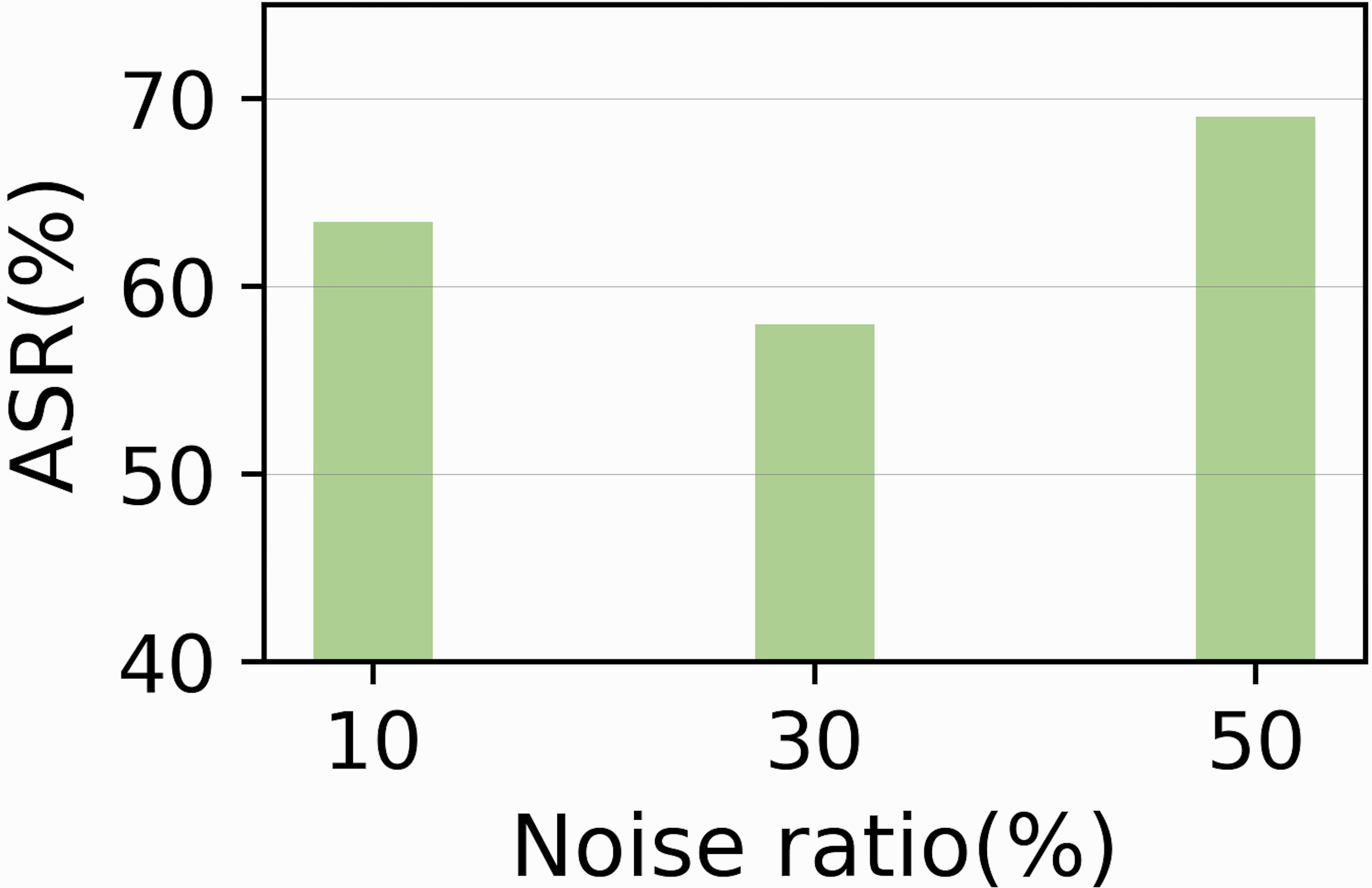}
    \label{fig:attack_fewrel}
    \end{minipage}} \hspace{2mm}
    \subfigure[TACRED]{ 
    \begin{minipage}[t]{0.46\linewidth}
    \centering    
    \includegraphics[scale=0.078]{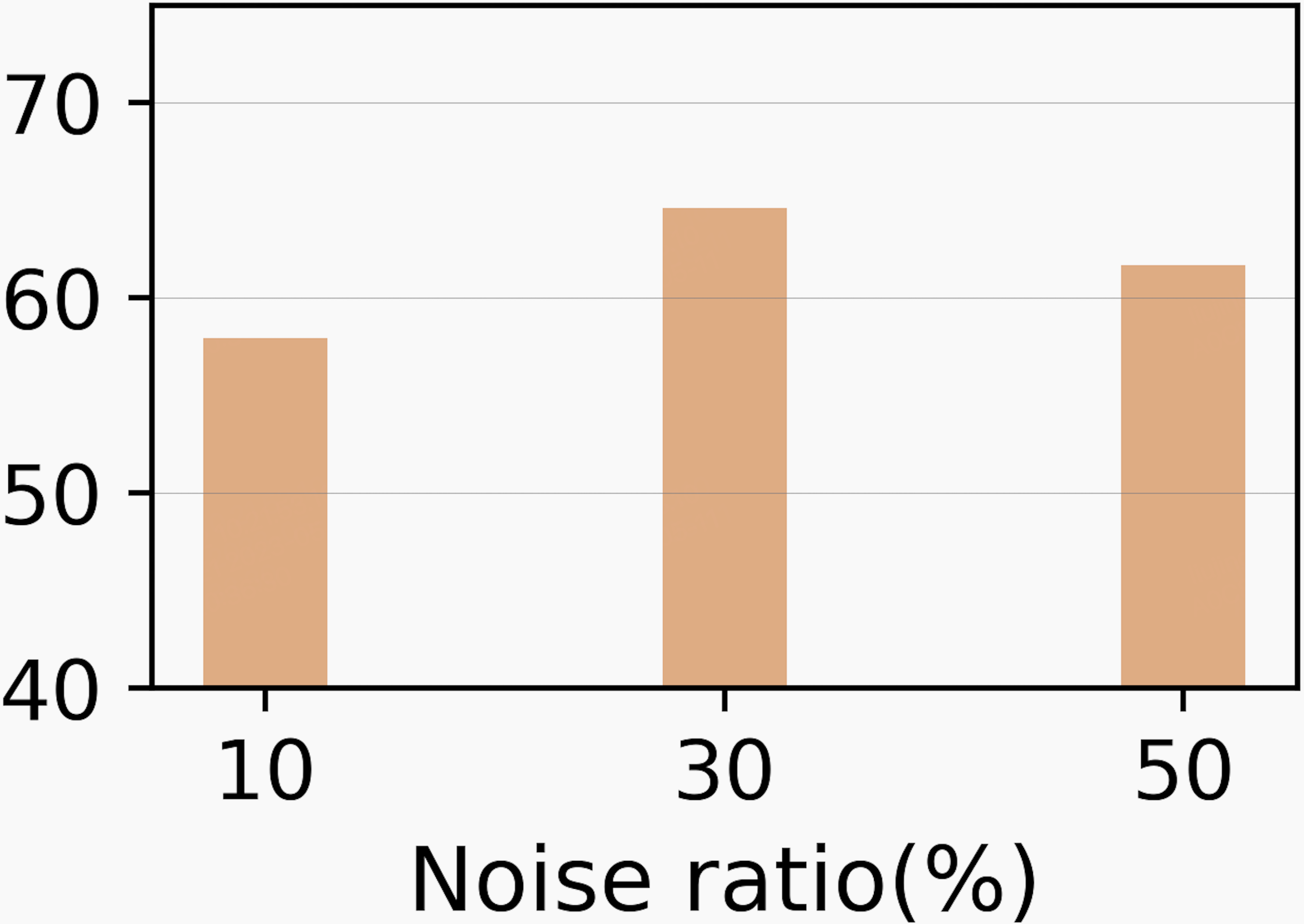}
    \label{fig:attack_tacred}
    \end{minipage}}
    \caption{\textbf{Attack success rate} with noise ratio of $\{10\%, 30\%, 50\%\}$.}
    
     \label{fig:attack}
\end{figure}

\begin{figure}[htbp]
    \centering
    \subfigure[30\% Noise Ratio]{   
    \begin{minipage}[t]{0.46\linewidth}
    \centering  
    \includegraphics[scale=0.08]{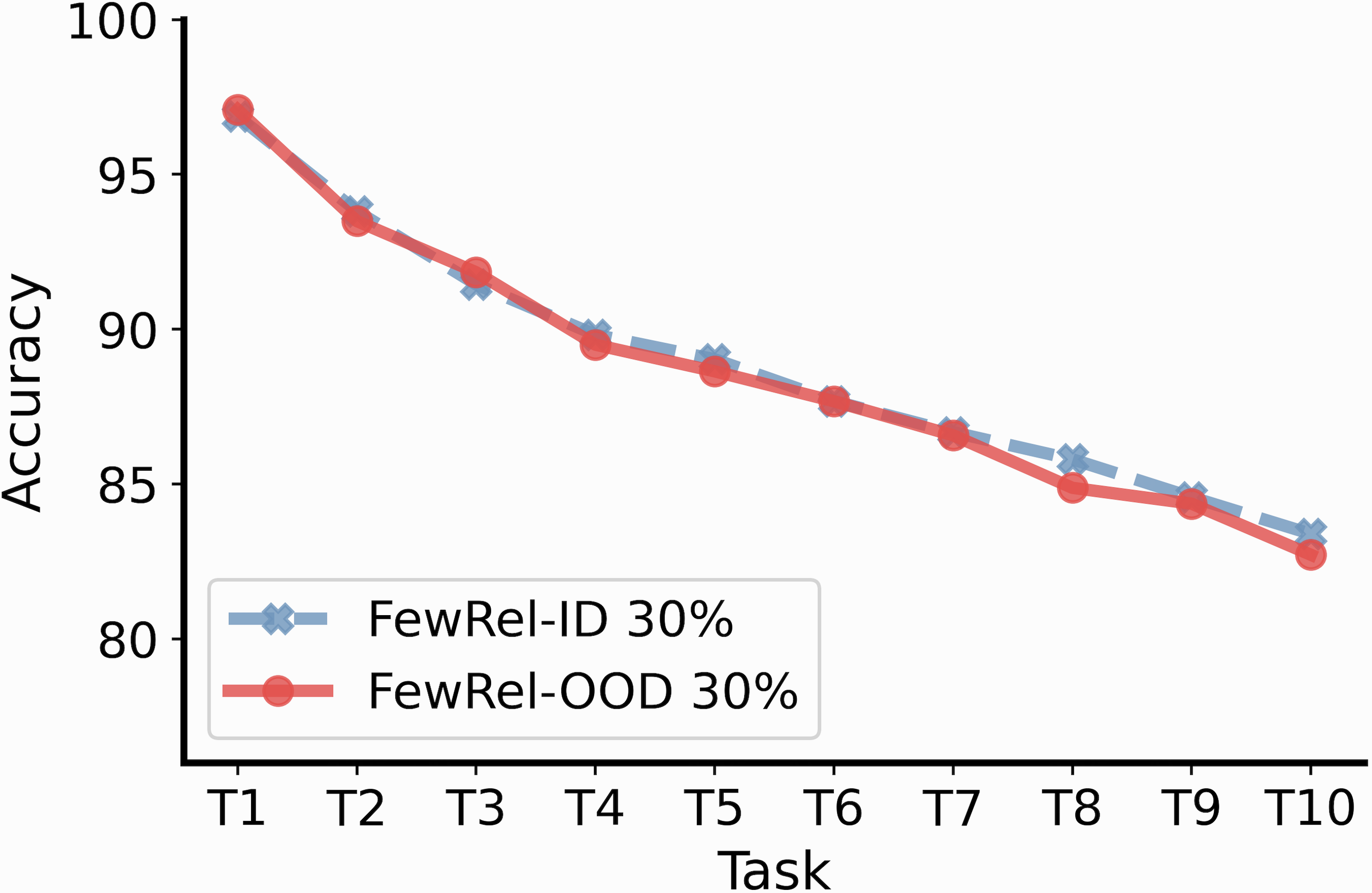}
    \label{fig:ood_fewrel}
    \end{minipage}}
    \subfigure[50\% Noise Ratio]{ 
    \begin{minipage}[t]{0.46\linewidth}
    \centering    
    \includegraphics[scale=0.08]{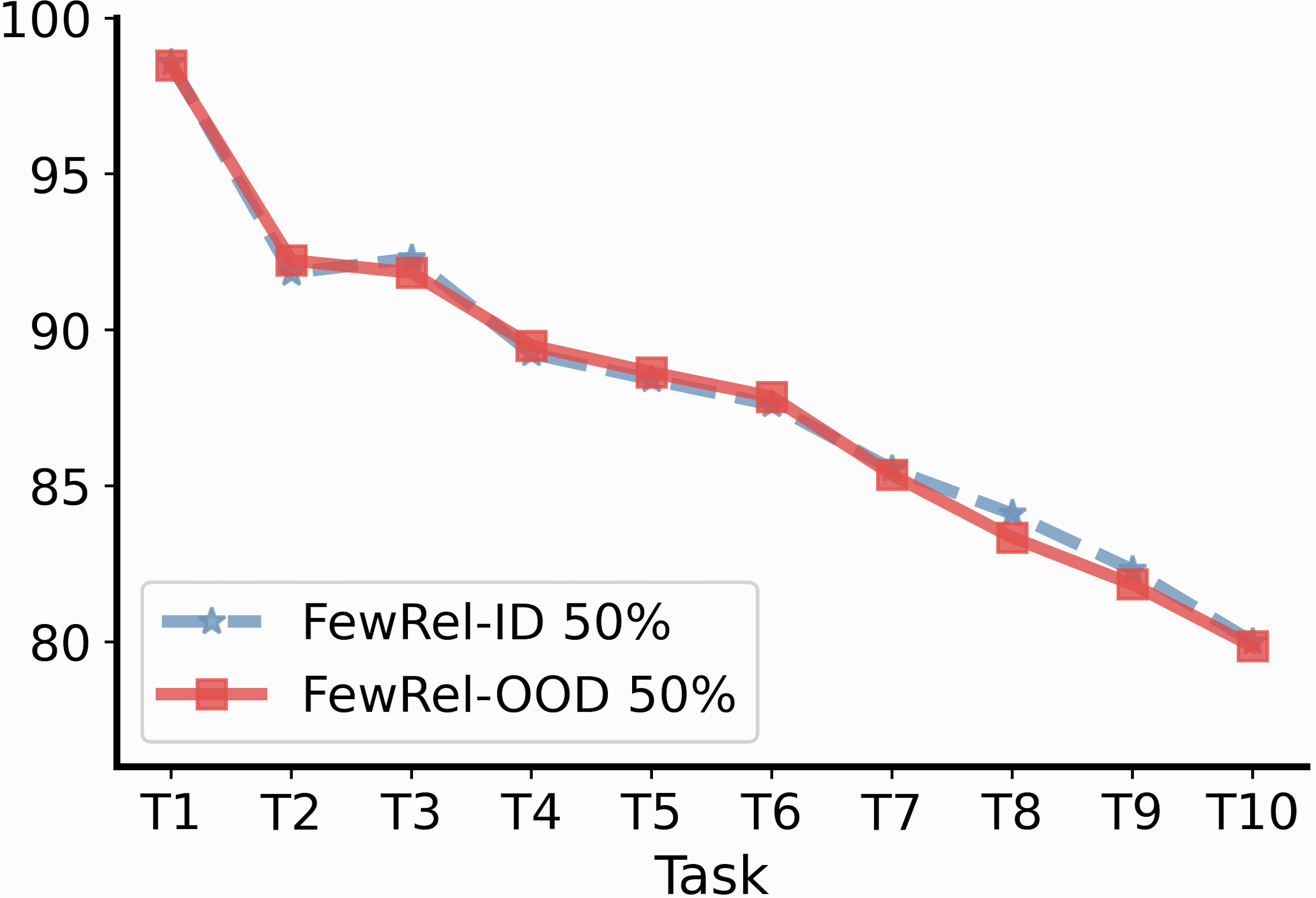}
    \label{fig:ood_tacred}
    \end{minipage}}
    \caption{\textbf{Accuracy}~($\%$) \textbf{on all seen relations} at the stage of learning current tasks with varying noise rates on FewRel ID set and OOD set~(TACRED).}
    
     \label{fig:ood_test}
\end{figure}

From the results in Table~\ref{tab:ablation_table}, we can conclude that compared with discarding the expected noisy samples directly, employing targeted adversarial attack can de facto make better use of the noisy ones, thus leading to performance improvements. To better investigate the influence of attack, we calculate attack success rate by Equation~\ref{eq:asr} on FewRel and TACRED with different noise rates. As shown in Figure~\ref{fig:attack}, by imposing a small perturbation on the input embedding, noise-guided attack can successfully force a great number of samples to the direction of their noisy labels in the feature space. 

\begin{table}[htbp]
    \centering
    \scalebox{0.8}{\begin{tabular}{cccccccc}\toprule
     \multicolumn{2}{c}{} &\multicolumn{3}{c}{FewRel} & \multicolumn{3}{c}{TACRED} \\
     \cmidrule(r){3-5} \cmidrule(r){6-8}
 \multicolumn{2}{c}{}& \multicolumn{3}{c}{$Acc$~($\%$)$\ \uparrow$} &\multicolumn{3}{c}{$Acc$~($\%$)$\ \uparrow$}\\
       Noise & Attack  & 10 & 30 & 50 & 10 & 30 & 50 \\\midrule
       \multicolumn{2}{c}{Discarding} &81.1 &80.7 &76.9 &77.8 &72.4 &68.5\\
      \checkmark & &83.0 &82.1 &78.0 &78.6 &75.5&70.5 \\
      \checkmark & \checkmark & \textbf{84.1} & \textbf{83.7} & \textbf{80.5} & \textbf{80.5} & \textbf{77.5} & \textbf{71.6} \\
 \bottomrule
    \end{tabular}}
    \caption{\textbf{Ablation studies} on the noise-guided attack, compared with noisy samples discarding.}
    \label{tab:ablation_table}
\end{table}

\subsection{Globally Open-set Label Noise}

In real-world applications, we expect a robust continual learner to be able to adapt well to noisy data streams, even with out-of-distribution~(OOD) samples. Empirical results have demonstrated that NaCL can successfully handle both closed-set label flips and open-set outliers. However, the meaning of \textit{open-set} we introduced before is only from a local perspective relative to the task progression. To explore the potential for noisy label learning from a global OOD set, as for FewRel, we further construct the label noise completely from TACRED. As the experimental results in Figure~\ref{fig:ood_test} show, NaCL achieves consistent performance when transferring from FewRel-ID to FewRel-OOD with varying noise rates, which demonstrates the superiority of NaCL for the strong noise resistance.

\section{Related Work}

\subsection{Continual Learning}
Prevalent methods for continual learning to tackle catastrophic forgetting problem can be categorized into three macro-types: \textit{rehearsal-based}, \textit{regularization-based}, and \textit{architecture-based} ones. Specifically, rehearsal-based methods construct a data buffer to save samples from older tasks to train with data at the current task~\cite{rebuffi-cvpr2017}. When the buffer storage is limited, exemplar selection techniques~\cite{Aljundi2019GradientBS} or generative modeling~\cite{sun2020lamal} are developed to help approximate the old data distribution. Viewed as exemplar-free methods without storing old task data, regularization-based ones consolidate old knowledge by limiting the learning rate on important parameters for previous tasks~\cite{doi:10.1073/pnas.1611835114}. Differently, architecture-based methods aim at having separate components for each task, and these task-specific components can be identified by expanding the network~\cite{loo2021generalized} or attending to task-specific sub-networks~\cite{pmlr-v162-gurbuz22a}.

Among them, rehearsal-based methods are substantiated to be the most effective paradigm in consolidating old knowledge~\cite{wang-etal-2019-sentence,sun2020lamal}. 
In this work, we consider combining NaCL with memory replay to help handle the severe forgetting problem. 

\subsection{Learning with Noisy Labels}
Deep neural networks are validated to easily overfit noisy labels resulting in poor generalization performance~\cite{pmlr-v70-arpit17a}. To improve model generalization with noisy labels, numerous approaches have been developed from various perspectives, \textit{e.g.}, loss correction~\cite{NEURIPS2018_ad554d8c}, robust loss functions with provable noise tolerance~\cite{pmlr-v119-ma20c}, sample-reweighting~\cite{ren18l2rw}, curriculum learning~\cite{zhou2021robust} and model co-teaching~\cite{han2018coteaching, pmlr-v97-yu19b}. The principle idea shared among these methods is to detect clean labels while discarding, down-weighting or relabeling the wrong labels.

Up to now, none of the works has focused on continual learning with noisy labels. Although strategies above seem to be well-handled for noisy labels, they are confined to \textit{closed-set} label flips and hence cannot be applied to our noisy-CRE setting. To be more generalized, our NaCL undertakes noise correction in the feature space to resolve both closed-set and open-set label noise.

\subsection{Contrastive Representation Learning}
As a dominant paradigm for representation learning, unsupervised contrastive learning~(UCL) has achieved comparable performance. The core idea behind UCL is to pull the anchor and the positive sample close to each other while pushing apart the anchor and the negative sample in embedding space~\cite{He2020MomentumCF}. Usually, the positives are produced from data augmentation while the negatives are random samples from the batch or the whole dataset. Concerned with the negative sampling distribution, recent works~\cite{robinson2021contrastive,NEURIPS2021_e5afb0f2} further validate that using \textit{hard negative samples}, i.e., the negative samples that are difficult to distinguish from the anchor can improve performance. Concurrently, supervised contrastive learning~(SCL) has developed to extend the unsupervised batch contrastive approach to a \textit{fully-supervised} setting that can leverage label information to select the positive and negative samples~\cite{NEURIPS2020_d89a66c7,gunel2021supervised}. 

Motivated by the hard-negative sampling strategies in UCL and the value of label information in SCL, our proposed NaCL utilizes both label information to retain the clean positives and attack the noisy samples to move closer to the decision boundary as a kind of hard negative mining.

\section{Conclusion}
Building on the recent wave of learning without forgetting, in this paper, we demonstrate current continual learners are vulnerable under natural label shifts. Hence, we propose a novel noise-resistant contrastive learning framework NaCL to correct the false contrastive pairs brought by the co-existence of closed-set and open-set label noise. Comprehensive experiments and analyses validate that our method can achieve the \textit{triple wins} that boost old knowledge, new task learning and noisy label robustness in one integrated algorithm.

\section*{Limitations}
The problem of natural shifts in label space over streaming data exists in various domains and datasets. To validate the effectiveness of our method for a better comparison, we conduct comprehensive experiments on relation extraction. Therefore, it is intriguing to generalize our noise-resistant contrastive learning framework to other applications for more robust continual learners. On the other hand, our method directly lineages the step of memory replay from previous work for its certified performance. However, from the perspective of efficiency and online learning, to maintain the plasticity-stability trade-off without replaying is worth further refinement. 

\section*{Ethics Statement}
There is an ongoing trend of developing continual learners to adapt the streaming data without forgetting previously learned knowledge. We hope our work can encourage the community to consider a more generalized setting of continual learning for better robustness. Moreover, our noise-resistant contrastive learning framework provides insight into dealing with false contrastive pairs with better views of positives and hard negatives mining.

\bibliography{emnlp2023}
\bibliographystyle{acl_natbib}

\clearpage
\appendix

\section{Supplementary Explanation}
\subsection{Real-world Noise}
\begin{table}[ht]
    \centering
    \begin{tabular}{cc}\toprule
    Dataset & Noise Level \\ \midrule
    Clothing1M & 38$\%$ ~\citep{JiahengWei2021LearningWN} \\
    Food-101N & 20$\%$~\citep{JiahengWei2021LearningWN}  \\
    NYT-10 & 35$\%$ ~\citep{li2020self}\\
    TACRED & 6.62$\%$~\citep{zhou2021learning} \\
    CoNLL03 & 5.38$\%$~\citep{zhou2021learning} \\
    Docred & 41.4$\%$~\citep{yao2019docred} \\
      \bottomrule
    \end{tabular}
    \caption{References for the noise level in Figure~\ref{fig:intro}.}
    \label{tab:reference}
\end{table}

\begin{table}[ht]
    \centering
    \begin{tabular}{cc}\toprule
    Notation & Meaning \\ \midrule
    $f_M$ & Main Model \\
    $\mathcal{E}_M$ & Main Feature Extractor \\
    Proj & Projector in Main Model \\
    $f_A$ & Auxiliary Model \\
    $\mathcal{E}_A$ & Auxiliary Feature Extractor \\
    $\mathcal{F}_A$ & Classifier in Auxiliary Model \\
    
      \bottomrule
    \end{tabular}
    \caption{Model Components Notation.}
    \label{tab:notation}
\end{table}

\newpage
\section{Training Algorithm} \label{appendix:algorithm}
We present the whole training procedure for $\mathcal{T}^k$ in Algorithm~\ref{alg:nacl}. 
\begin{algorithm}[ht]
\begin{algorithmic}[1] 
    \caption{Training procedure for $\mathcal{T}^k$ }
    \Ensure{$\mathcal{D}_{\rm train}^k$: contaminated training set of the $k$-th task, $f_M(\cdot,\theta)$: main model, $f_A(\cdot,\theta^\ast)$: auxiliary model, $\mathcal{B}$: memory buffer with exemplars stored}
    
    \Require{learning rate $\eta$ for $f_M$ and $f_A$,  batch size $m_s$, training epochs $E_1, E_2$, perturbation radius $\epsilon$, noise-guided attack step $s$}

    \For{epoch$=1,\cdots, E_2$} \textcolor{orange}{\Comment{Selection}}
    \State Sample a batch $\{(x_i,y_i)\}^{m_s}_{i=1}$ from $\mathcal{D}_{\rm train}^k$
    \State Training $f_A$ by Equation~\ref{eq:crossentropy}
    \EndFor
    \State Obtain $\widetilde{\mathcal{D}}_{\rm clean}$ and $\widetilde{\mathcal{D}}_{\rm noisy}$ by Equation~\ref{eq:selection}  
    
    \For{ $(x_i,y_i) \in \widetilde{\mathcal{D}}_{\rm noisy}$} \textcolor{orange}{\Comment{Attack}}
    \State $x^\prime_i \leftarrow x_i + \delta$, where $\delta\sim$ Uniform$(-\epsilon, \epsilon)$
    \For{ \textit{fixed} step $s=1, \cdots, S$}
    \State  Perform noise-guided attack by Equation~\ref{eq:attack}
    \EndFor 
    \State Group $(x_i,y_i)$ with success attack to $\mathcal{D}_{\rm att\mbox{-}pos}$ and $\mathcal{D}_{\rm neg}$ otherwise \quad 
    \EndFor 

    \For{epoch$=1,\cdots, E_1$} \textcolor{orange}{\Comment{$\mathcal{T}^k$ Training}}
    \State Sample a batch $\{(x_i,y_i)\}_{i=1}^{m_s}$ from $\widetilde{\mathcal{D}}_{\rm clean}$
    \State Contrastive training of $f_M$ by Equation~\ref{eq:nacl}
    \EndFor

    \If{$\mathcal{T}^k$ is not the first task}\textcolor{orange}{\Comment{Replay}}
    \State Update memory buffer $\mathcal{B}$ with exemplars selected from $\widetilde{\mathcal{D}}_{\rm clean}$ 
    \For{epoch$=1,\cdots, E_1$} 
    \State Sample a batch $\{(x_i,y_i)\}_{i=1}^{m_s}$ from $\mathcal{B}$
    \State Training $f_M$ by Equation~\ref{eq:scl}
    \EndFor
    \EndIf

\label{alg:nacl}
\end{algorithmic}
\end{algorithm}

\section{Hyper-parameter Setup}
All the hyper-parameters in our experiments for reproduction are shown in Table~\ref{tab:hyperparameters}. 
\begin{table*}[t]
\renewcommand{\arraystretch}{1.3}
    \centering
    \begin{tabular}{cccc}\toprule
    Parameter & Meaning & FewRel & TACRED \\ \midrule
    \multirow{2}{*}{$\gamma$} & \multirow{2}{*}{selection threshold~(Equation~\ref{eq:selection})} & {0.8,0.6,0.5} & {0.9,0.75,0.6 } \\
     &  & {for $\{10\%,30\%, 50\%\}$} & {for $\{10\%,30\%, 50\%\}$}  \\
    $\lambda$ & trade-off for attack~(Equation~\ref{eq:attack}) & 0.1 & 0.1 \\
    $\epsilon$ & perturbation size~(Equation~\ref{eq:attack}) & 0.1 & 0.1\\
    $s$ & attack steps~(Equation~\ref{eq:attack}) & 5 & 5\\
    $\tau$ & temperature~(Equation~\ref{eq:nacl}) & \multicolumn{2}{c}{0.1,0.05,0.2 for $\{10\%,30\%, 50\%\}$} \\ 
     $n$ & total task numbers & 10 & 10 \\
    $\mathcal{C}$ & classes of each incremental task & 8 & 4 \\
     $\eta$ & learning rate for $f_M$ and $f_A$ & 1e-5 & 2e-5 \\
     $m_s$ & training batch size & 16 & 16 \\
     $dim$ & projection dimension & 64 & 64 \\
     $E_1$ & training epoch of $f_M$ for new relations & 1 & 1\\
     $E_2$ & training epoch of $f_A$ for selection & 3 & 3\\
      \bottomrule
    \end{tabular}
    \caption{List of hyper-parameters for our approach to reproduce the results in Table~\ref{tab: main_res}.}
    \label{tab:hyperparameters}
\end{table*}

\end{document}